\documentclass[letterpaper]{article} 
\usepackage{aaai24}
\usepackage{times}  
\usepackage{helvet}  
\usepackage{courier}  
\usepackage[hyphens]{url}  
\usepackage{graphicx} 
\usepackage{array}
\usepackage{enumitem}
\usepackage{natbib}  
\usepackage{caption} 
\usepackage{algorithm}
\usepackage{algorithmic}
\usepackage{subcaption}
\usepackage[export]{adjustbox}
\usepackage{tcolorbox}
\usepackage{scalerel}
\usepackage{multirow}
\usepackage{times}
\usepackage{latexsym}
\usepackage{graphicx}
\usepackage{algorithm}
\usepackage{algorithmic}
\usepackage{amssymb}
\usepackage{tikz,lipsum}
\usepackage{booktabs} 
\usepackage{amsmath,amsfonts}
\usepackage[export]{adjustbox}
\usepackage{tcolorbox}
\usepackage{enumitem}
\setlist{leftmargin=2mm}
\usepackage{pifont}
\usepackage{xcolor}
\usepackage{listings}

\DeclareCaptionStyle{ruled}{labelfont=normalfont,labelsep=colon,strut=off} 
\title{\textit{CLIPSyntel}: CLIP and LLM Synergy for Multimodal Question Summarization in Healthcare}
\author {
    Akash Ghosh\textsuperscript{\rm 1}\equalcontrib,
    Arkadeep Acharya\textsuperscript{\rm 1}\equalcontrib,
    Raghav Jain\textsuperscript{\rm 1},
    Sriparna Saha\textsuperscript{\rm 1},
    Aman Chadha\textsuperscript{\rm 2,3}\footnote{Work does not relate to position at Amazon.},\\
    Setu Sinha\textsuperscript{\rm 4}
}
\affiliations {
    \textsuperscript{\rm 1}Indian Institute of Technology Patna,
    \textsuperscript{\rm 2}Stanford University,
    \textsuperscript{\rm 3}Amazon AI,
    \textsuperscript{\rm 4}Indira Gandhi Institute of Medical Sciences\\
    \texttt{\{akash\_2321cs19, arkadeep\_2101ai41, sriparna\}@iitp.ac.in, raghavjain106@gmail.com, hi@aman.ai,
    drsinhasetu@gmail.com}
}

\usepackage{bibentry}

\begin{document}

\maketitle

\begin{abstract}
In the era of modern healthcare, swiftly generating medical question summaries is crucial for informed and timely patient care. Despite the increasing complexity and volume of medical data, existing studies have focused solely on text-based summarization, neglecting the integration of visual information. Recognizing the untapped potential of combining textual queries with visual representations of medical conditions, we introduce the Multimodal Medical Question Summarization (MMQS) Dataset. This dataset, a major contribution of our work, pairs medical queries with visual aids, facilitating a richer and more nuanced understanding of patient needs. We also propose a framework, utilizing the power of Contrastive Language Image Pretraining(CLIP) and Large Language Models(LLMs), consisting of four modules that identify medical disorders, generate relevant context, filter medical concepts, and craft visually aware summaries.  Our comprehensive framework harnesses the power of CLIP, a multimodal foundation model, and various general-purpose  LLMs, comprising four main modules: the medical disorder identification module, the relevant context generation module, the context filtration module for distilling relevant medical concepts and knowledge, and finally, a general-purpose LLM to generate visually aware medical question summaries. Leveraging our MMQS dataset, we showcase how visual cues from images enhance the generation of medically nuanced summaries. This multimodal approach not only enhances the decision-making process in healthcare but also fosters a more nuanced understanding of patient queries, laying the groundwork for future research in personalized and responsive medical care.\\%
\textbf{Disclaimer}: The article features graphic medical imagery, a result of the subject's inherent requirements.

\end{abstract}

\section{Introduction}
In the midst of the global COVID-19 pandemic, healthcare systems have been inundated with an unprecedented volume of inquiries, concerns, and uncertainties. Individuals worldwide are posing health-related inquiries on online platforms, seeking clarity and guidance on symptoms, prevention, treatments, and vaccinations. These questions often employ everyday language and encompass both pertinent and extraneous details that may not directly pertain to the sought-after solutions. The complexity and urgency of the situation, coupled with a considerable imbalance in the doctor-to-patient ratio across many countries, have made the ability to swiftly and comprehensively comprehend a patient's query paramount  \cite{abacha2019summarization,yadav2021reinforcement}. In this context, medical question summarization has emerged as a vital tool to distill information from consumer health questions, ensuring the provision of accurate and timely responses. However, existing works have overlooked the untapped potential of integrating visual data, particularly images, with textual information. The motivations for focusing on visual aids within medical question summarization (MQS) are manifold. A significant portion of the population lacks familiarity with medical terms needed to accurately describe various symptoms, and some symptoms are inherently challenging to articulate through text alone. Patients may also be confused between closely related symptoms, such as distinguishing between skin dryness and skin rash. The combination of text and images in medical question summarization can offer enhanced accuracy and efficiency, providing a richer context that textual analysis alone may miss. This approach recognizes the complex nature of patient queries, where photographs of symptoms, medical reports, or other visual aids could provide crucial insights. By focusing on the integration of images, researchers and healthcare providers can respond to the evolving challenges of modern healthcare communication.\par
\begin{figure*}[hbt]
	\centering
	\includegraphics[width=0.90\linewidth]{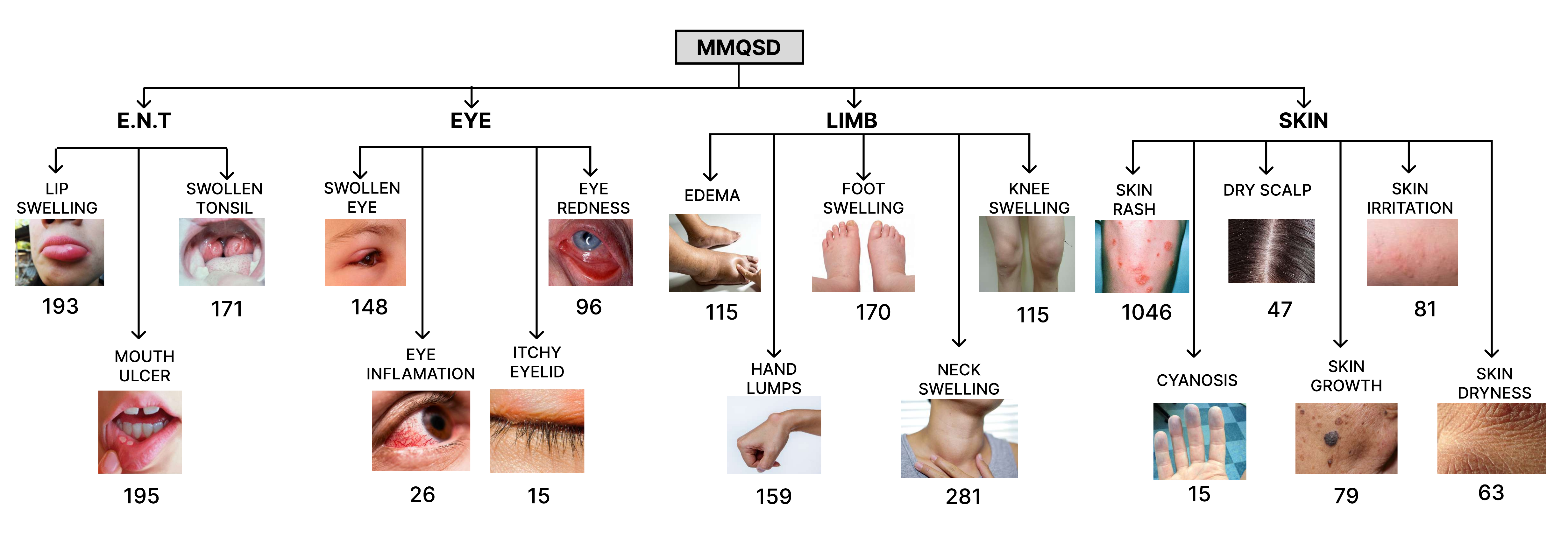}
  \setlength{\abovecaptionskip}{0pt}
	\caption {Broad categorization of medical disorders in the MMQS Dataset (MMQSD). The number of data points corresponding to each category has been provided under each category in the above figure. }
	\label{fig:ann}
\end{figure*} 
Large Language Models (LLMs) \cite{kojima2023large} and Vision Language Models (VLMs) \cite{zhang2023vision} have exhibited remarkable capacities in generating human-like text and multimedia content. This prowess has driven their deployment across the medical domain, predominantly for domain-specific tasks such as chest radiograph summarization \cite{thawkar2023xraygpt} and COVID-19 CT report generation \cite{liu2021medical}. Yet, their application in multimodal medical question summarization remains uncharted territory. Leveraging zero-shot and few-shot learning capabilities of these models \cite{dong2022survey} offers a compelling advantage, especially for tasks like multimodal medical question summarization, characterized by inherent data scarcity. However, while the potential of LLMs and VLMs in this domain is undeniable, they aren't without constraints. Predominantly, generic LLMs and VLMs often lack a solid grounding in task-specific knowledge, risking the generation of summaries that might miss intricate details like symptoms, diagnostic tests, and medical complexities. On the visual front, even as VLMs have excelled in typical visual-linguistic tasks, medical imaging presents unique challenges. Efforts like SkinGPT4 \cite{zhou2023skingpt}, which fine-tunes MiniGPT4 \cite{zhu2023minigpt} on skin disease images and clinical notes, are still notably domain-specific. Medical images are inherently complex, demanding a profound understanding of medical terminology and visual conventions, often necessitating an expert medical practitioner's perspective for accurate interpretation. This complexity, combined with potential gaps in contextual understanding, can result in models producing misleading or irrelevant summaries.\par
To address the limitations of LLMs and VLMs in multimodal medical question summarization, we've conceived the \textbf{\textit{CLIPSyntel}} framework. The first stage, the Medical Disorder Identification Module, combines Large Language Model(LLM) with Contrastive Language Image Pretraining(CLIP) \cite{radford2021learning}, forming a novel zero-shot classification approach to identify disorders from visual images, utilizing the unique benefits of zero or few-shot learning. Subsequently, during the Context Generation Phase, LLM adds context to the medical query, focusing on vital components like symptoms, tests, and procedures, mindful of the potential pitfalls of extraneous content. Addressing this challenge, the Context Filtration Module filters the content using a multimodal knowledge selection technique, preserving only the most relevant information. The Summary Generation Phase follows, where an LLM crafts the final medically accurate summaries based on the distilled knowledge. Finally, to validate \textit{CLIPSyntel}, we created the \textbf{\textit{MMQS dataset}}, rich in visual and textual representations of patients' symptoms and queries\footnote{\url{https://github.com/AkashGhosh/CLIPSyntel-AAAI2024}}. This careful, step-by-step structuring of \textit{CLIPSyntel} allows it to bridge the divide between general-purpose models and the niche requirements of medical question summarization, effectively integrating textual and visual data to create precise and context-aware medical summaries. To summarize we make the following main contributions:\\
\textbf{A novel task} of Multimodal Medical Question Summarization for generating medically nuanced summaries.\\
\textbf{A novel  dataset}, \textit{MMQS} Dataset, to advance the research in this area.\\
\textbf{A novel metric} \textit{MMFCM} to quantify how well the model captures the multimodal information in the generated summary.
\\
\textbf{A novel framework}, \textit{ClipSyntel} that harnesses the power of CLIP and LLMs to augment the patient question with an additional rich context from visual symptoms for the generation of final summaries.

\begin{figure*}[htbp]
  \centering
  \begin{subfigure}{0.45\textwidth}
    \includegraphics[width=0.75\linewidth,center]{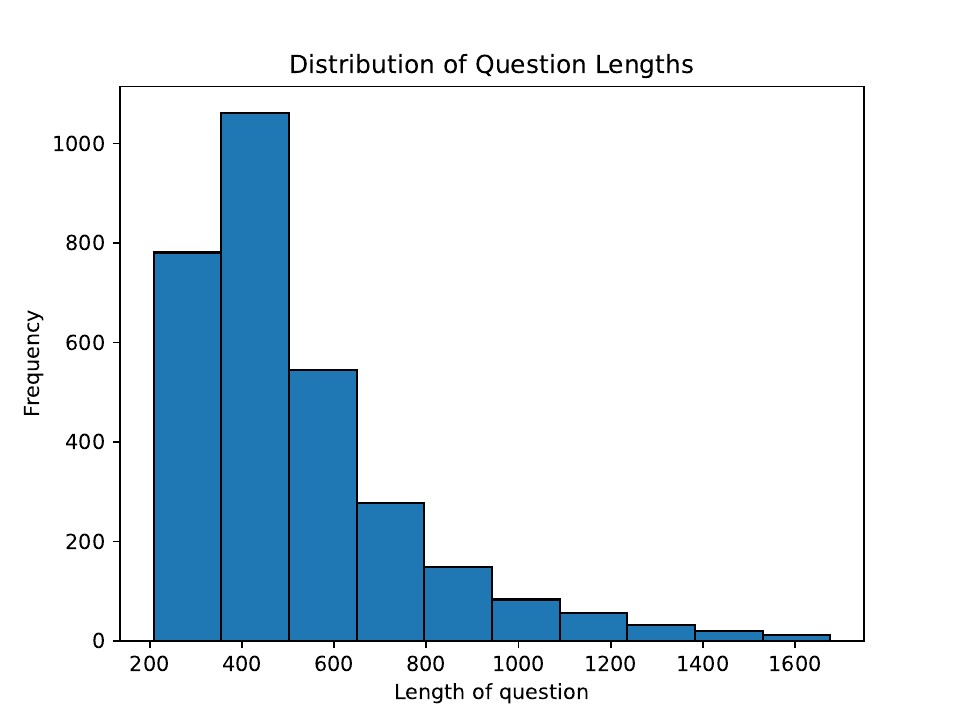}
    \caption{Distribution of Question Lengths.}
    \label{fig:text_length_distribution}
  \end{subfigure}
  \hfill
  \begin{subfigure}{0.45\textwidth}
    \includegraphics[width=0.75\linewidth,center]{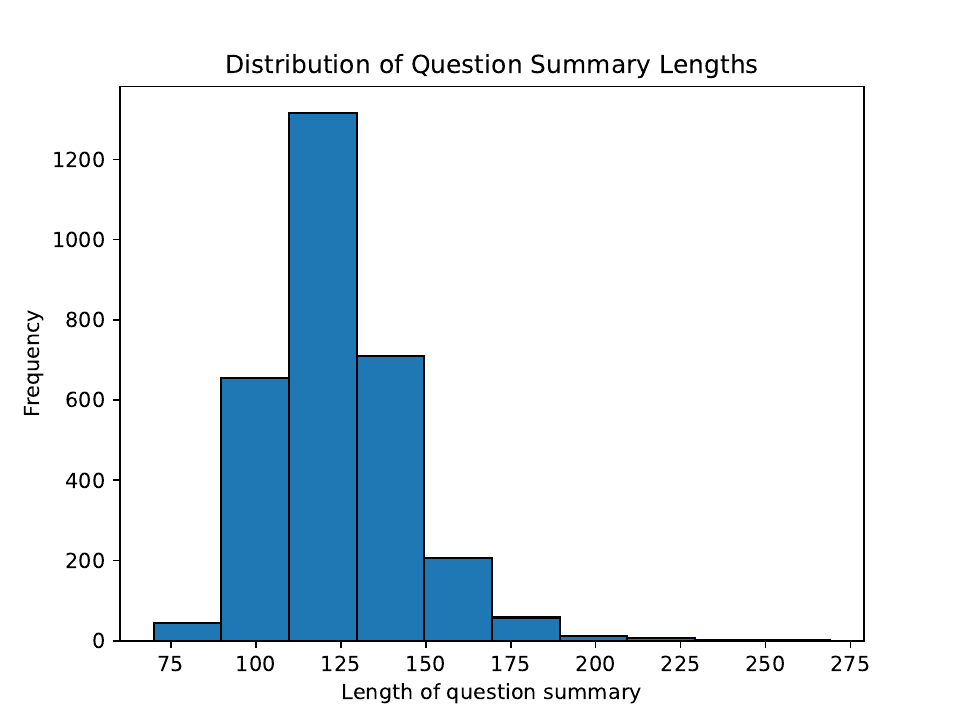}
    \caption{Distribution of Summary Lengths.}
    \label{fig:Question_summ_length_disb}
  \end{subfigure}
  \hfill
  \caption{Distribution of question lengths and their summaries in the MMQS dataset; }
  \label{fig:Statistics}
\end{figure*}
\section{Related Works}
\textbf{Medical Question Summarization:} The task of Medical Question Summarization (MQS) was initially introduced in 2019 with the development of the MeQSum dataset \cite{abacha2019summarization} specifically for this purpose. Early research on MQS employed vanilla seq2seq models and pointer generator networks to generate summaries. In 2021, a contest was held focusing on the generation of medical domain summaries \cite{abacha2021overview}. Participants utilized various pre-trained models such as PEGASUS \cite{zhang2020pegasus}, ProphetNet \cite{qi2020prophetnet}, and BART \cite{lewis2019bart}. Techniques like multi-task learning were employed, utilizing BART to jointly optimize question summarization and entailment tasks \cite{mrini2021joint}. Another approach involved reinforcement learning with question-aware semantic rewards derived from two subtasks: question focus recognition (QFR) and question type identification (QTR) \cite{yadav2021reinforcement}.

\textbf{Role of Multimodality:} In order to receive accurate treatment and guidance from a medical expert, our responsibility is to effectively and efficiently communicate the medical symptoms which can be done with additional visual cues. Previously, many studies also showed that adding multimodality through visual cues improves the performance of various medical tasks. \citet{tiwari2022dr} shows how multimodal information helps in building better Disease Diagnosis Virtual Assistants. \citet{delbrouck-etal-2021-qiai} also showed how incorporating images helps in better summarization of radiology reports. \citet{gupta2022dataset} also showed the benefit of incorporating videos for medical QA task
The current work is motivated by this idea of how integrating the patient's provision of an image of their medical condition alongside the text, improves the generation of more medically rich summaries. We curate a multimodal dataset based on an existing MQS dataset, focusing on a predefined set of symptoms, and propose a novel framework that combines cutting-edge Multimodal Foundation Models like CLIP with Large Language Models (LLMs). To our best understanding, this work is the first to tackle the task of question summarization in the medical domain, particularly within a multimodal setting.
\section{MMQS (Multimodal Medical Question Summarization) Dataset}
\vspace{1mm}
To the best of our knowledge, a freely available multimodal question summarization dataset that includes both textual questions and corresponding medical images of patients' conditions does not currently exist. To construct such a dataset, we utilized the preexisting HealthCareMagic Dataset, which is derived from the MedDialog data introduced by \cite{mrini2021joint}. The initial dataset comprised 226,395 samples, of which 523 were found to be duplicates. To ensure data integrity and fairness, we eliminated these duplicate entries. In order to ascertain the medical symptoms or signs that could effectively be conveyed through visual means, we consulted a medical professional who also happens to be a co-author of this paper. Following a series of brainstorming sessions and carefully analyzing the dataset, we identified 18 symptoms that are hard to specify only through text. These 18 symptoms are broadly divided into four broad groups of multimodal symptoms to incorporate into our dataset. These groups (refer to figure \ref{fig:ann}) are categorized as ENT (Ear, Nose, and Throat), EYE (Eye-related), LIMB (Limbs-related), and SKIN (Skin-related). From the ENT category, we selected symptoms including lip swelling, mouth ulcers, and swollen tonsils. From the EYE category, we choose swollen eyes, eye redness, and itchy eyelids. For the LIMB category, we included symptoms such as edema, foot swelling, knee swelling, hand lumps, and neck swelling. Lastly, from the SKIN category, we included symptoms of skin rash, skin irritation, and skin growth. To extract images for the corresponding symptoms, Bing Image Search API\footnote{https://www.microsoft.com/en-us/bing/apis/bing-image-search-api} was used. Then the images extracted are verified by a group of medical students who are led by a medical expert. Therefore, in curating our ultimate multimodal dataset, we selectively included instances from the HealthCareMagic dataset that featured textual mentions of body parts such as skin, eyes, ears, and others, both within the questions and their corresponding summaries. We refrained from directly searching for symptom names, as our intention was to encompass instances from the dataset where uncertainty regarding the medical condition exists. Consequently, patients implicitly allude to symptoms rather than explicitly stating their names. To identify these relevant samples, we employed the Python library FlashText\footnote{https://pypi.org/project/flashtext/1.0/} for efficient term matching in the aforementioned dataset. The Python library Textblob4 was used to correct the spelling of many improperly spelled words. To address the misspelling of numerous words, the Textblob4\footnote{https://pypi.org/project/textblob/0.9.0/} Python library was employed. Following an initial exploration using FlashText, we initially collected approximately 5000 samples. After meticulous manual validation of each sample, our refinement process led us to a final dataset comprising 3015 samples where multimodality could be incorporated for the final dataset creation.

\begin{figure*}
\centering
\begin{subfigure}{.492\textwidth}
  \centering
  \includegraphics[width=1\linewidth]{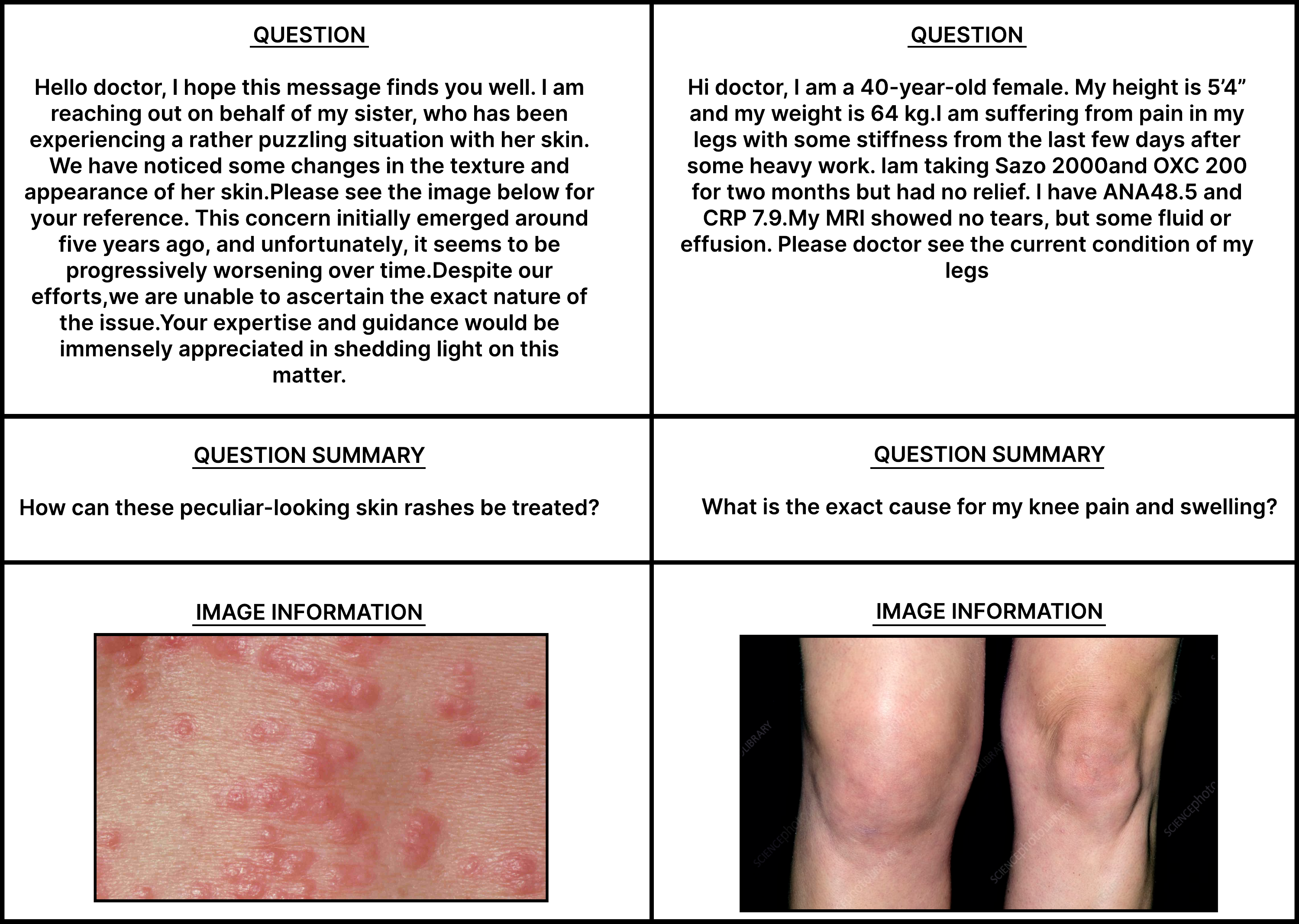}
  \caption{Sample instances from the MMQS-dataset.}
  \label{fig:sample}
\end{subfigure}\hfill 
\begin{subfigure}{.492\textwidth}
  \centering
  \vstretch{1.192}{\includegraphics[width=1.04\linewidth,center]{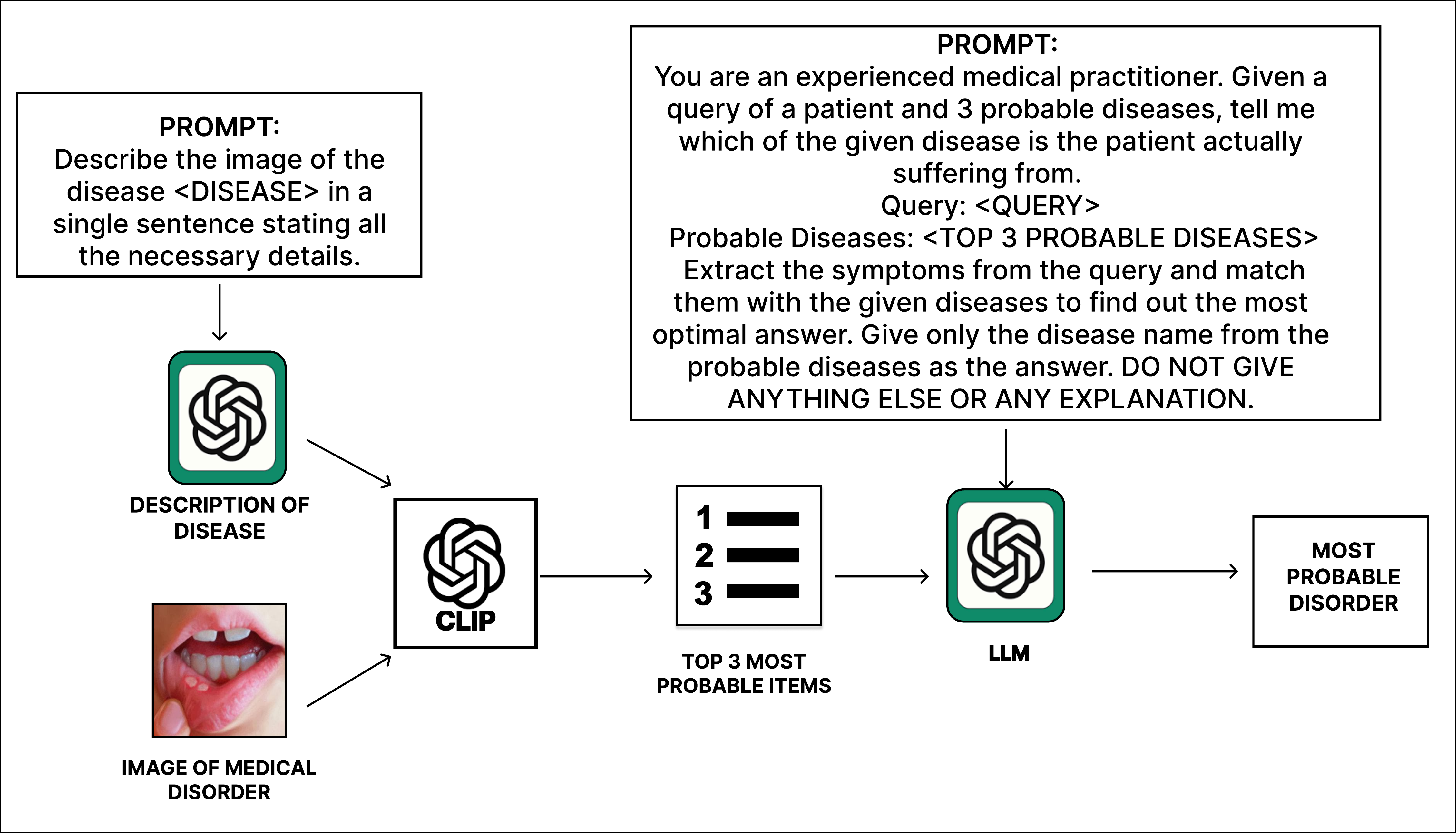}}
  \caption{Medical Disorder Identification Module}
  \label{fig:sub2}
\end{subfigure}
\caption{Comprehensive illustration of Sample Instances from the MMQS-dataset and Medical Disorder Identification Module.}
\label{fig:test}
\end{figure*}

\begin{figure*}[hbt]
	\centering
	{\includegraphics[width=1\linewidth, center]{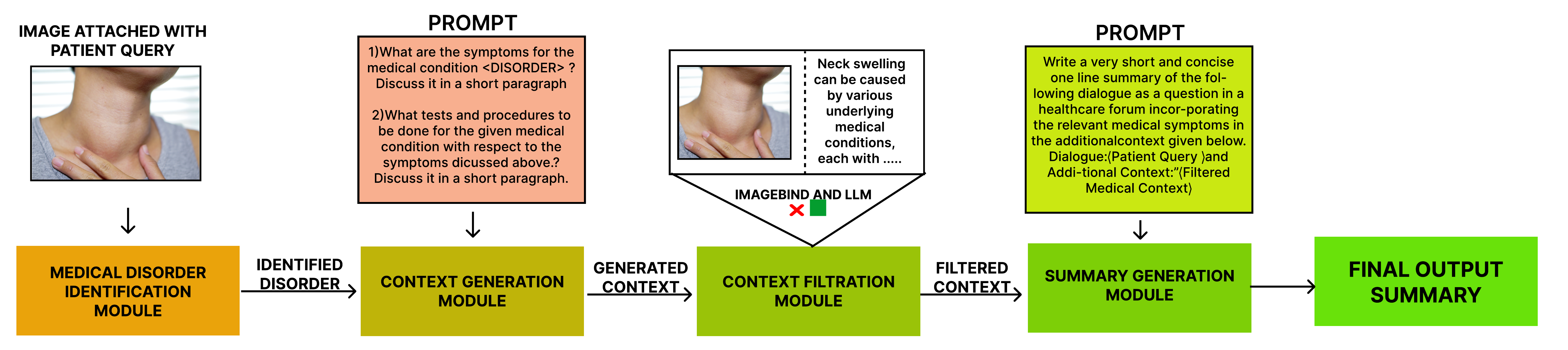}}
  \setlength{\abovecaptionskip}{0pt}
	\caption {Flowchart depicting the workflow of the proposed model \textit{CLIPSyntel}. The diagram illustrates the step-by-step process of \textit{ClipSyntel}'s functioning.}
	\label{fig:model}
\end{figure*} 
\subsection{Data Annotation}
We extracted 100 random samples from the filtered dataset, each comprising a patient's inquiry alongside its corresponding summary, and gave them to the medical expert for annotation. The methodology through which the medical expert annotated these samples is explained with the given example. For example, suppose the patient is complaining regarding something that has happened near their tonsils, but they are not able to name the exact medical disorder they are suffering from. But their textual reference gave enough context to understand that the patient is suffering from the disorder named \textbf{swollen-tonsils}. After understanding this, the medical expert adds statements like \textbf{\textit{Please see what happened to my tonsils in the image below}} in the context and accordingly adds a visual image representation of the disorder. In this way, visual medical signs are incorporated into the textual medical question. Additionally, the medical expert deduced that the conventional golden summaries did not align well with the multimodal queries, prompting the need for corresponding updates. To address these issues, the medical expert himself modified these 100 samples by revising the questions, incorporating multimodal information, and adapting the golden summaries accordingly. Subsequently, we engaged with medical students and graduates to annotate the remaining samples based on the guidelines given by the medical expert\footnote{The students were compensated through gift vouchers and honorarium.}. To quantify the level of agreement among annotators, we calculated the kappa coefficient (k), which yielded a value of 0.78. This coefficient indicates a noteworthy level of consistent annotation. Following this rigorous process, the Multimodal Medical Question Summarization (MMQS) dataset comprised a total of 3,015 samples. The median length of the question in the final dataset is 442 words and the median length of the question summary is 121 words, and example data points are also shown in Table-\ref{fig:sample}. The distributions of question length and question summary length are shown in Figure \ref{fig:text_length_distribution} and Figure \ref{fig:Question_summ_length_disb}, respectively.

\section{Methodology}
\textbf{Problem Formulation:} Each data point consists of a patient query in textual form $T$ and an image $I$ of the medical disorder that the patient is suffering from or trying to address in his textual query. The final output is a concise summary $S$ of the query which incorporates information from both the modalities. This section presents the novel architecture of \textit{\textbf{CLIPSyntel}} shown in Figure-\ref{fig:model}, a multimodal, knowledge-grounded framework designed to generate medically nuanced summaries.For a comprehensive understanding, we partition our approach into four distinct modules: (i) \textbf{Medical Disorder Identification Module}, (ii) \textbf{Contextual Information Generation Module}, (iii) \textbf{Context Filtration Module}, and (iv) \textbf{Summary Generation Module}

\subsection{Medical Disorder Identification}
For the process of generating pertinent medical knowledge it is necessary to accurately identifying the medical disorder that the patient is experiencing. To achieve this, we employed the following structured approach: \textbf{(1) Utilization of CLIP:} CLIP, designed to establish connections between images and text, is initially applied. Despite its applicability to various visual classification tasks, CLIP's effectiveness in our Medical Disorder Identification Task using only the names of the disorder is limited due to the complex nature of medical images. \textbf{ (2) Enhanced Contextualization through Prompts:} To provide better context for a particular medical disorder, we prompted GPT 3.5 using the prompt: \textbf{Describe the image of the disease $\{DISEASE\}$ in a single sentence stating all the necessary details}. This approach yields improved contextual information for the specific medical condition. \textbf{(3) Identification of Probable Disorders:} When presented with an image of a medical disorder, we provided the contextual information for all 18 medical disorders, along with the corresponding image, to the CLIP model. Subsequently, we selected the top 3 most likely medical disorders based on CLIP's analysis. \textbf{(4) Final Prediction with GPT-3.5:} These three probable diseases, along with the medical query, is then subsequently passed to LLM(GPT-3.5) using the prompt which is shown in Figure-\ref{fig:sub2} to predict the final medical disorder. Considering only the most probable disorder prediction from CLIP yields an accuracy of 84 \%. However, when incorporating the top 3 most probable disorders alongside the context after passing through the LLM, the accuracy is enhanced to 87\% in a zero-shot setup. The entire pipeline for this module, displaying its logical flow and connections, is depicted in Figure \ref{fig:sub2}.

\subsection{Contextual Medical Knowledge Generation}
In cases where patients may lack awareness of their medical condition, and textual inquiries are insufficient, acquiring additional knowledge about their specific \textit{\textbf{symptoms}} and the \textit{\textbf{necessary tests and procedures}} (as advised by medical experts) proves essential for creating a meaningful summary, especially in a multimodal setup. Formally, after identification of the medical disorder $M$, we formulate a set of prompts $P=\{p_0, p_1, \ldots, p_n\}$ with the aim of generating diverse contextual information about that disorder $M$, $KS=\{ks_0, ks_1, \ldots, ks_n\}$ by promting \textit{\textbf{GPT-3.5}} as follows. The Prompts used to generate contextual information are as below:\par
\textbf{\textit{ (1) What are the symptoms of the medical condition }\textlangle \textbf{medical disorder} }\textrangle?\par
 \textbf{\textit{(2) What tests and procedures need to be done for the medical condition \textlangle \textbf{medical disorder}}} \textrangle?

\begin{table*}[]
\centering
\scalebox{0.62}{%
\begin{tabular}{c|c|ccc|cccc|c|}
\cline{2-10}
 &
  \multirow{2}{*}{\textbf{Model}} &
  \multicolumn{3}{c|}{\textbf{ROUGE}} &
  \multicolumn{4}{c|}{\textbf{BLEU}} &
   \multirow{2}{*}{\textbf{BERTScore}} \\ 
   \cline{3-9}  &     & \textbf{R1}    & \textbf{R2}   & \textbf{RL}    & \textbf{B1}    & \textbf{B2}    & \textbf{B3}    & \textbf{B4}    &   \\ \hline
\multicolumn{1}{|c|}{\multirow{3}{*}{\rotatebox[origin=c]{90}{\textbf{LLAMA-2}}}}  & LLM(only texual query)&  0.188 & 0.051 & 0.159 & 0.144 & 0.064 & 0.032 & 0.016 & 0.822 
  \\
\multicolumn{1}{|c|}{}   & & & & & & & & &                       \\
\multicolumn{1}{|c|}{} & Clip+GPT-3.5 & 0.189 & 0.052 & 0.161 & 0.156 & 0.070 & 0.036 & 0.018 & 0.819                         \\
\multicolumn{1}{|c|}{}   & & & & & & & & &                       \\
\multicolumn{1}{|c|}{}   &  \textit{CLIPSyntel} & \textbf{0.249} & \textbf{0.081} & \textbf{0.211} & \textbf{0.195} & \textbf{0.102} & \textbf{0.059} & \textbf{0.034} & \textbf{0.864}  \\ \hline \hline
\multicolumn{1}{|c|}{\multirow{4}{*}{\rotatebox[origin=c]{90}{\textbf{RedPajama}}}} & LLM(only texual query) & 0.146 & 0.023 & 0.125 & 0.086  & 0.030 &  0.011 & 0.004  & 0.828
  \\
\multicolumn{1}{|c|}{} &&&&&&&&&                      \\
\multicolumn{1}{|c|}{} & Clip+GPT-3.5 & 0.148 & 0.025 & 0.132 & 0.087 & 0.0314 & 0.012 & 0.004 & 0.831                     \\
\multicolumn{1}{|c|}{}   & & & & & & & & &                         \\
\multicolumn{1}{|c|}{}                      &  \textit{CLIPSyntel} &  \textbf{0.157} & \textbf{0.0279} & \textbf{0.140} & \textbf{0.088} & \textbf{0.032} & \textbf{0.014} & \textbf{0.006} & \textbf{0.838} \\ \hline \hline
\multicolumn{1}{|c|}{\multirow{4}{*}{\rotatebox[origin=c]{90}{\textbf{Vicuna}}}} &  LLM(only textual query)& 0.372 & 0.160 & 0.318 & 0.385 & 0.243 & 0.161 & 0.102 & 0.905 \\
\multicolumn{1}{|c|}{}    & & & & & & & & &     
\\\multicolumn{1}{|c|}{} & Clip+GPT-3.5 & 0.384 & 0.164 & 0.321 & 0.391 & 0.245 & 0.167 & 0.105 & 0.906                     \\
\multicolumn{1}{|c|}{}    & & & & & & & & &                     \\
\multicolumn{1}{|c|}{}   & \textit{CLIPSyntel} & \textbf{0.391} & \textbf{0.167} &  \textbf{0.33} & \textbf{0.40}  & \textbf{0.25} & \textbf{0.171} & \textbf{0.108} & \textbf{0.910} \\ \hline \hline
\multicolumn{1}{|c|}{\multirow{4}{*}{\rotatebox[origin=c]{90}{\textbf{FLAN-T5}}}} & LLM(only textual query) & 0.220 & 0.042 & 0.205 &  0.129 & 0.052 & 0.026 & 0.012 & 0.88 \\
\multicolumn{1}{|c|}{}   & & & & & & & & &                       \\
\multicolumn{1}{|c|}{} & Clip+ GPT-3.5 & 0.220 & 0.042 & 0.205 & 0.136  &  0.056 & 0.026 & 0.012 & 0.88                     \\
\multicolumn{1}{|c|}{}   & & & & & & & & &                      \\
\multicolumn{1}{|c|}{}   & \textit{CLIPSyntel} & \textbf{0.243} &  \textbf{0.068} & \textbf{0.215} & \textbf{0.182} & \textbf{0.096} & \textbf{0.052} & \textbf{0.0256}  &  \textbf{0.891}
\\ \hline \hline
\multicolumn{1}{|c|}{\multirow{4}{*}{\rotatebox[origin=c]{90}{\textbf{GPT-3.5}}}} & LLM(only textual query)  & 0.451 & 0.227 & 0.396 & 0.471 & 0.330 & 0.243 & 0.170 & 0.920
  \\

\multicolumn{1}{|c|}{} & & & & & & & & &                      \\
\multicolumn{1}{|c|}{} & Clip + GPT-3.5 & 0.452 & 0.226 & 0.399 & 0.471 & 0.331 & 0.242 & 0.171 & 0.920                      \\
\multicolumn{1}{|c|}{} & & & & & & & & &                       \\
\multicolumn{1}{|c|}{}                      &  \textit{CLIPSyntel}      & \textbf{0.463} & \textbf{0.241} & \textbf{0.409} & \textbf{0.478} & \textbf{0.342} & \textbf{0.257} & \textbf{0.186} & \textbf{0.921} \\ \hline
\end{tabular}%
   }
\caption{Performance of various \textit{CLIPSyntel} models and corresponding baselines, evaluated using automatic metrics with different  LLMs.}
\label{tab:quant}
\end{table*}
\subsection{Multimodal Medical Knowledge Filtration}
While we anticipate that \textbf{GPT-3.5} will generate useful and pertinent information based on the given prompt, it occasionally engages in hallucinations and produces irrelevant content in relation to the primary inquiry. This hallucinated text has the potential to divert Large Language Models (LLMs) from generating high-quality summaries. It may also generate content that is not pertinent for doctors to investigate the case swiftly. Particularly in fields such as medicine, one must exercise heightened caution to minimize misinformation, given that errors could have serious consequences for patients. To tackle this issue, we present a filtering strategy called Multimodal Medical Knowledge Filtration, which operates as follows. Formally, for each $ks_i$, which corresponds to a specific text field in the $KS$, it is further broken down into a set of $k$ sentences $ks_i=\{s_1, s_2, \ldots, s_m\}$ using sentence tokenization\footnote{\url{https://www.nltk.org/api/nltk.tokenize.html}}. To achieve this, we employ a pre-trained multimodal model, ImageBind \cite{girdhar2023imagebind}, which serves as an off-the-shelf encoder-based multimodal model $enc(\cdot)$ that maps a token sequence (text) and an image to their respective feature vectors embedded in a unified vector space. Cosine similarity $sim(\cdot, \cdot)$ is then used to measure the relevance of each sentence to the image. Formally, a sentence $s_j$ is retained if $sim(enc(s_j), enc(I)) > Th$, where $I$ is the image associated with the medical disorder, and $Th$ is the predefined similarity threshold. Building upon this, we define the subset of retained knowledge sentences, having undergone the Multimodal Medical Knowledge Filtration module (\textit{MMKFS}), as follows:

\vspace{-6mm}

\begin{equation}
\begin{aligned}
\resizebox{0.89\hsize}{!}{$
    Ms'_i = \{s_j \,|\, s_j \in ks_i \,\text{and } \text{sim}(\text{enc}(s_j), \text{enc}(I)) > Th\}
$}
\end{aligned}
\end{equation}

\vspace{-1mm}

With this approach, the filtered knowledge sentences $MS'=\{ms'_0, \ldots, ms'_n\}$ not only hold a high degree of visual-textual alignment with the corresponding medical disorder image but are also contextually relevant.

\subsection{Summary Generation Module}
This component is crafted to leverage the refined knowledge sentences, denoted as $MS'$ in order to craft informative and contextually relevant summaries for patients' healthcare inquiries. The Summary Generation Module functions with a pre-trained Large Language Model (LLM), referred to as $LM$. For summary creation, $LM$ is provided a specially crafted prompt $P{in}$ that integrates both the actual patient query $X$ and the generated medical disorder knowledge $MS'$. The outcome of this process is the generated summary text, denoted as $S$. In a more formal representation, this process can be articulated as: $S = LM(P{in}(X, MS'))$.
The prompt used to generate the final summary is presented below:\par
\textbf{\textit{Write a very short and concise one line summary of the following dialogue as a question in a healthcare forum incorporating the relevant medical symptoms in the additional context given below.
 Dialogue:\textlangle Patient Query \textrangle  and Additional Context:\textlangle Filtered Medical Context\textrangle}}

\begin{table}[]
\centering
\scalebox{0.55}{%
\renewcommand{\arraystretch}{1.78}
\begin{tabular}{cc|cccc|}
\cline{2-6}
\multicolumn{1}{c|}{}                          & \textbf{Model(GPT-3.5)}     & \textbf{Clinical-EvalScore} & \textbf{Factual Recall} & \textbf{Omission Rate} & \textbf{MMFCM Score} \\ \hline
\multicolumn{1}{|c|}{\multirow{6}{*}}    &  LLM(Patient Query)       &  3.4       &     0.748     &   0.235             &   0.9                \\

\multicolumn{1}{|c|}{} & Clip + GPT-3.5 &  \textbf{3.51}       &    \textbf{0.792}      &    \textbf{0.2215}            &  \textbf{1.52}               \\

\multicolumn{1}{|c|}{}                         & \textit{CLIPSyntel} &  \textbf{3.62}       &    \textbf{0.818}      &    \textbf{0.2217}            &  \textbf{1.6}               \\

\hline \hline

\multicolumn{2}{|c|}{\textbf{Annotated Summary}}               &   \textbf{4.1}     & \textbf{0.889}        & \textbf{0.144}               &   \textbf{1.86}              \\ \hline
\end{tabular}%
}
\caption{Human evaluation scores of the best \textit{CLIPSyntel} model and their corresponding baselines across different metrics.}
\label{tab:human}
\end{table}

\section{Experimental Results and Discussion}
\vspace{1mm}

\textbf{Experimental Setup: }We leveraged the following general purpose LLMs for summary generation module: RedPajama\footnote{https://huggingface.co/togethercomputer/RedPajama-INCITE-Chat-3B-v1}, FLAN-T5 ~\cite{chung2022scaling}, Vicuna \cite{zheng2023judging} and GPT-3.5. We tested these LLMs in different settings: With only the patient's question(text) in the prompt, With the patient's question combined with the knowledge obtained from  (CLIP+GPT3.5), and then our proposed framework (\textit(CLIPSyntel)). We have set the similarity threshold parameter $Th$ to 0.5 and temperature to 0.5 across all settings based on a thorough investigation. We utilize ROUGE ~\cite{lin2004rouge}, BLEU~\cite{papineni2002bleu}, and BERTScore~\cite{zhang2019bertscore} as automatic evaluation metrics. For the purpose of human evaluation, we collaborate with a  medical expert and a few medical students. We have identified four distinct and medically nuanced metrics for this evaluation: clinical evaluation score, factual recall \cite{abacha2023investigation}, omission rate\cite{abacha2023investigation}, and our newly introduced MMFCM score.\par
\textbf{Automated Evaluation:}
The results presented in Table \ref{tab:quant} provide valuable insights into the performance of various models, highlighting the effectiveness of different approaches in the context of the task. Below, we discuss some key observations and trends: \textbf{(1) \textit{CLIPSyntel} Performance:} Across all LLMs, \textit{CLIPSyntel} consistently performed the best. This indicates the robustness and versatility of \textit{CLIPSyntel}, demonstrating its ability to outshine other base models. The results suggest the capability of \textit{CLIPSyntel}'s design to effectively leverage both textual and visual information. These results also show that adding contextual information does help in our task. \textbf{(2) GPT-3.5 based Models' Superiority:} Among the various models evaluated, the GPT-3.5 based models stood out as the top performers compared to open-source models. The high scores across multiple metrics reflect the sophisticated design and capability of GPT-3.5. \textbf{(3) Best Performance Among Open-Source LLMs:} Vicuna was the best performer among open-source LLMs. Its scores were notably higher compared to other open-source counterparts, signaling its potential as an effective alternative for specialized tasks.\par 
\textbf{Human Evaluation:}
The human evaluation was done by a team of medical students led by a doctor. The team was given 10 \% of the dataset (selected at random) for evaluation purpose and was asked to rate the summaries generated, which takes only patient question (text) as context, patient query in addition to the medical disorder context (CLIP + GPT-3.5) and finally the summary generated by our proposed pipeline (\textit{CLIPSyntel}). The following metrics are used for the evaluation: \textbf{(1) Clinical Evaluation Score:} The doctor and his team were asked to rate the summaries between 1 (poor) and 5 (good) based on their overall relevance, consistency, fluency, and coherence. \textbf{(2) Multi-modal fact capturing metric (MMFCM):}
We propose a new metric to evaluate how well a model incorporates relevant medical facts and identifies the correct disorder in a multimodal setup. MMFCM is calculated by considering both the facts extracted from the medical query and image and assessing whether they are correctly incorporated in the generated summary. The metric accounts for (1) The ratio of correct facts in the summary to the total number of relevant medical facts. (2) Additional scores based on the accuracy of the disorder's detection, with values ranging from +2 for fully correct to -1 for incorrect identification. See Algorithm 1 for the detailed algorithm. 

\begin{algorithm}
	\caption{MMFCM Method }
	\begin{algorithmic}
\REQUIRE{$F_{m} = \{ \text{fact}_{m,1}, \text{fact}_{m,2}, \ldots, \text{fact}_{m,n-1}, \text{fact}_{m,n} \}$}
\COMMENT{Relevant Medical facts from query 'm'.}

        \REQUIRE{$Sf_{m}$ =$\{{Summfact_{m,1},\dots,Summfact_{m,n}}\} $}\\
        \COMMENT{ Relevant Medical facts of summary of query 'm'.}
        
\STATE $\#CorrectFacts_{m} = \lvert F_{m} \cap Sf_m \rvert$ \newline
\COMMENT{Number of correct medical facts in each summary}

\IF{$\left(\text{Correct Medical Disorder phrase} \in \{ F_{m} \cap Sf_m \} \right)$}
    \STATE $\#CorrectFacts_{m} += 2$
\ELSIF{$ \left( \text{Partially correct disorder phrase} \in \{ F_{m} \cap Sf_m \} \right)$}
    \STATE $\#CorrectFacts_{m} += 1$
\ELSIF{$ \left( \text{Incorrect disorder phrase} \in \{ F_{m} \cap Sf_m \} \right)$}
    \STATE $\#CorrectFacts_{m} += -1$
\ELSE
    \STATE $\#CorrectFacts_{m} += 0$
\ENDIF

\RETURN MMFCM = $\frac{\#\text{CorrectFacts}_{m}}{\lvert F_{m} \cap Sf_m \rvert}$
\end{algorithmic}
\end{algorithm}

\textbf{(3) Medical Fact Based Metrics:} We employed the Factual Recall and Omission Recall metrics \cite{abacha2023investigation} to assess the extent to which the generated summary captures medical facts compared to the gold standard annotated summary.\par
Table \ref{tab:human} offers a comprehensive comparative analysis, underscoring \textit{ClipSyntel}'s distinct advantages over various baseline methods across multiple human evaluation performance metrics. Notably, our proposed architecture excels the summaries that incorporate only the patient's textual information, highlighting the valuable impact of fusing visual symptom cues with textual patient queries, resulting in more clinically nuanced summaries. Moreover, the augmentation of medical knowledge is affirmed by substantial enhancements in Factual Recall and Omission Rate metrics, further corroborated by our proposed Multimodal fact Capturing Metric(MMFCM) score. Overall, \textit{ClipSyntel} emerges as a clear frontrunner in human evaluation metrics, outperforming competing baselines by a considerable margin.\par
\begin{figure}[]
	\centering
	\includegraphics[width=0.9\linewidth, center]{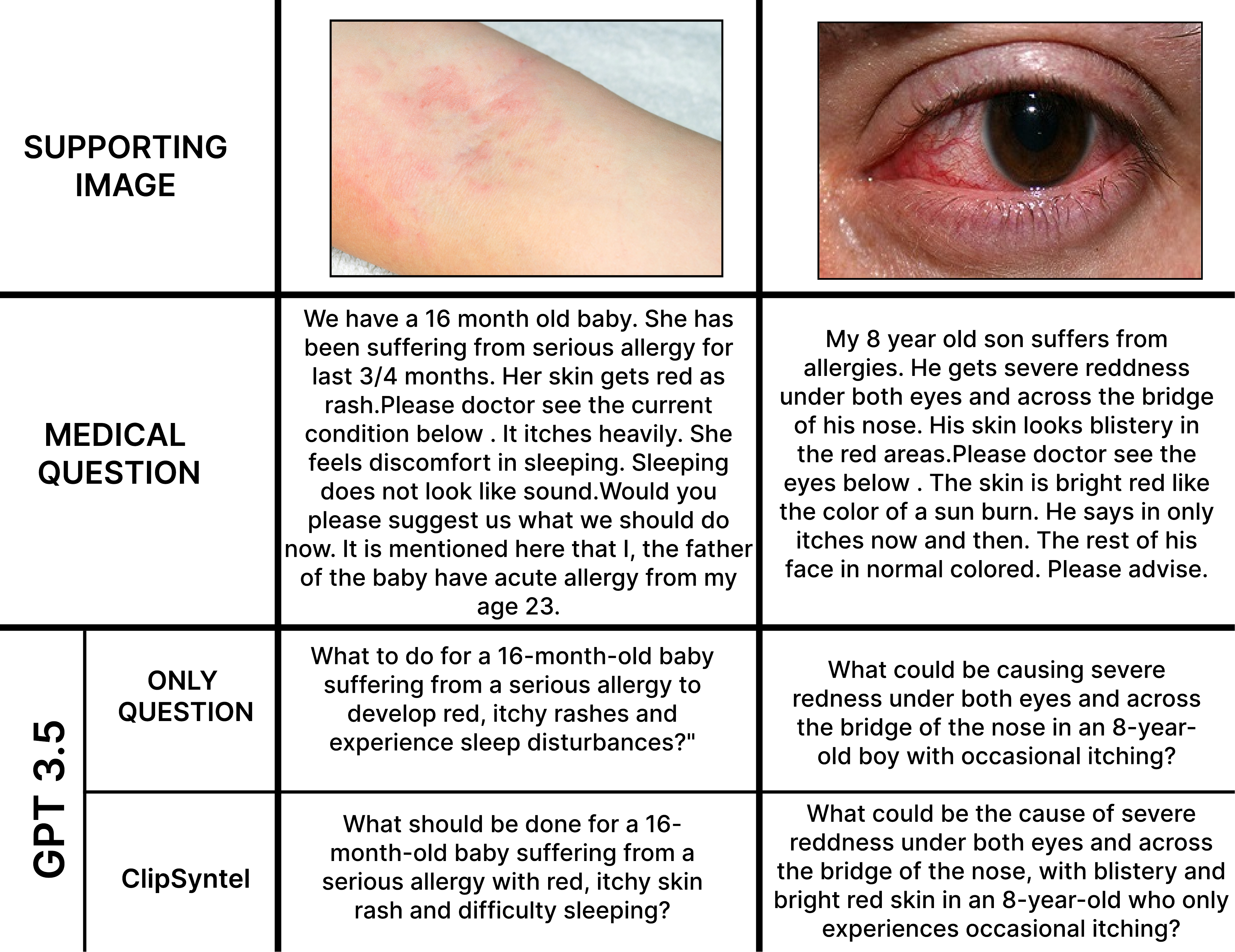}
  \setlength{\abovecaptionskip}{0pt}
	\caption {Sample Summaries generated by two variations GPT-3.5 model: only question and \textit{CLIPSyntel}}
	\label{fig:qualt}
\end{figure} 
\textbf{Qualitative analysis: }The Figure \ref{fig:qualt} provides an insightful comparison of the summaries generated by two variations of the GPT-3.5 model, namely the "Only Question" and "\textit{CLIPSyntel}" configurations, for medical inquiries. The analysis mainly focuses on two aspects: (1) \textit{CLIPSyntel} Captures Medical Disorders Correctly: The \textit{CLIPSyntel} variant demonstrates a noteworthy ability to identify and represent medical conditions with precision. In both cases presented in the table, \textit{CLIPSyntel} has shown an understanding of the underlying medical issues. (2) \textit{CLIPSyntel} Does Less Hallucinations:  \textit{CLIPSyntel}'s summaries stick to the information explicitly provided in the medical questions. Unlike the "Only Question" variant that might introduce slight changes or uncertainties, \textit{CLIPSyntel} maintains a strict adherence to the facts. This ability to avoid unnecessary extrapolations or inaccuracies enhances \textit{CLIPSyntel}'s reliability. 
 
\section{Risk Analysis}
It's important to note several limitations in our approach. 
We restricted our study to a zero-shot prompting strategy, not fully examining how prompt variations might affect results. Second, although CLIP performs well in low-resource environments, we must be cautious of potential misclassifications of medical images, as these could lead to serious or even fatal misinformation in a summary generation. Despite the promising performance of \textit{CLIPSyntel} in various scenarios, the risk of misinformation in healthcare is significant. Therefore, involving a medical expert is vital to ensure that our AI model serves as an aid, not a replacement, for medical professionals.
\section{Conclusion and Future Work}
In this study, we delve into the impact of incorporating multi-modal cues, particularly visual information, on question summarization within the realm of healthcare. We present the MMQS dataset, comprising 3015 multimodal medical queries with golden summaries that merge visual and textual data. This novel collection fosters new assessment techniques in healthcare question summarization. We also introduce the \textit{CLIPSyntel} framework, leveraging LLMs and the CLIP model, to enhance summaries with visual symptom details. CLIP enables symptom classification, and an ImageBind-filtering module mitigates content hallucination. In our future endeavors, we aspire to develop a Vision-Language model capable of extracting the intensity and duration details of symptoms and integrating them into the patient query's final summary generation. Furthermore, our expansion plans encompass incorporating medical videos and addressing scenarios involving code-mixed patient queries.

\section{Ethical Considerations}
Summarization in healthcare necessitates strong ethical considerations, particularly regarding safety, privacy, and potential bias. To address these concerns in our project with the MMQS dataset, we implemented several proactive measures. We collaborated closely with medical professionals  and also obtained IRB approval to ensure ethical rigor and patient privacy. We rigorously followed legal and ethical guidelines \footnote{\url{https://www.wma.net/what-we-do/medical-ethics/declaration-of-helsinki/}} during dataset validation, integration of images, and annotation of summaries. Medical experts were engaged throughout the process, providing validation and correction of the dataset and also validating the outputs of the models. The proposed dataset is based on original HealthCareMagic
Dataset; the medical questions/samples are taken from this dataset. The incorporation of multimodality into the task is done under the full supervision of a medical professional. Additionally, we ensured user privacy by not disclosing identities.

\section{Acknowledgement}
Akash Ghosh expresses his heartfelt gratitude to the Technology Innovation Hub (TIH) IIT Patna and SERB for the awarded fellowship, a crucial source of support that significantly bolsters his research endeavors.

\bibliography{anonymous-submission-latex-2024}

\begin{thebibliography}{25}
\providecommand{\natexlab}[1]{#1}

\bibitem[{Abacha and Demner-Fushman(2019)}]{abacha2019summarization}
Abacha, A.~B.; and Demner-Fushman, D. 2019.
\newblock On the summarization of consumer health questions.
\newblock In \emph{Proceedings of the 57th Annual Meeting of the Association
  for Computational Linguistics}, 2228--2234.

\bibitem[{Abacha et~al.(2021)Abacha, M’rabet, Zhang, Shivade, Langlotz, and
  Demner-Fushman}]{abacha2021overview}
Abacha, A.~B.; M’rabet, Y.; Zhang, Y.; Shivade, C.; Langlotz, C.; and
  Demner-Fushman, D. 2021.
\newblock Overview of the MEDIQA 2021 shared task on summarization in the
  medical domain.
\newblock In \emph{Proceedings of the 20th Workshop on Biomedical Language
  Processing}, 74--85.

\bibitem[{Abacha et~al.(2023)Abacha, Yim, Michalopoulos, and
  Lin}]{abacha2023investigation}
Abacha, A.~B.; Yim, W.-w.; Michalopoulos, G.; and Lin, T. 2023.
\newblock An Investigation of Evaluation Metrics for Automated Medical Note
  Generation.
\newblock \emph{arXiv preprint arXiv:2305.17364}.

\bibitem[{Chung et~al.(2022)Chung, Hou, Longpre, Zoph, Tay, Fedus, Li, Wang,
  Dehghani, Brahma et~al.}]{chung2022scaling}
Chung, H.~W.; Hou, L.; Longpre, S.; Zoph, B.; Tay, Y.; Fedus, W.; Li, E.; Wang,
  X.; Dehghani, M.; Brahma, S.; et~al. 2022.
\newblock Scaling instruction-finetuned language models.
\newblock \emph{arXiv preprint arXiv:2210.11416}.

\bibitem[{Delbrouck, Zhang, and Rubin(2021)}]{delbrouck-etal-2021-qiai}
Delbrouck, J.-B.; Zhang, C.; and Rubin, D. 2021.
\newblock {QIAI} at {MEDIQA} 2021: Multimodal Radiology Report Summarization.
\newblock In \emph{Proceedings of the 20th Workshop on Biomedical Language
  Processing}, 285--290. Online: Association for Computational Linguistics.

\bibitem[{Dong et~al.(2022)Dong, Li, Dai, Zheng, Wu, Chang, Sun, Xu, and
  Sui}]{dong2022survey}
Dong, Q.; Li, L.; Dai, D.; Zheng, C.; Wu, Z.; Chang, B.; Sun, X.; Xu, J.; and
  Sui, Z. 2022.
\newblock A survey for in-context learning.
\newblock \emph{arXiv preprint arXiv:2301.00234}.

\bibitem[{Girdhar et~al.(2023)Girdhar, El-Nouby, Liu, Singh, Alwala, Joulin,
  and Misra}]{girdhar2023imagebind}
Girdhar, R.; El-Nouby, A.; Liu, Z.; Singh, M.; Alwala, K.~V.; Joulin, A.; and
  Misra, I. 2023.
\newblock Imagebind: One embedding space to bind them all.
\newblock In \emph{Proceedings of the IEEE/CVF Conference on Computer Vision
  and Pattern Recognition}, 15180--15190.

\bibitem[{Gupta, Attal, and Demner-Fushman(2022)}]{gupta2022dataset}
Gupta, D.; Attal, K.; and Demner-Fushman, D. 2022.
\newblock A Dataset for Medical Instructional Video Classification and Question
  Answering.
\newblock arXiv:2201.12888.

\bibitem[{Kojima et~al.(2023)Kojima, Gu, Reid, Matsuo, and
  Iwasawa}]{kojima2023large}
Kojima, T.; Gu, S.; Reid, M.; Matsuo, Y.; and Iwasawa, Y. 2023.
\newblock Large language models are zero-shot reasoners. arXiv.

\bibitem[{Lewis et~al.(2019)Lewis, Liu, Goyal, Ghazvininejad, Mohamed, Levy,
  Stoyanov, and Zettlemoyer}]{lewis2019bart}
Lewis, M.; Liu, Y.; Goyal, N.; Ghazvininejad, M.; Mohamed, A.; Levy, O.;
  Stoyanov, V.; and Zettlemoyer, L. 2019.
\newblock Bart: Denoising sequence-to-sequence pre-training for natural
  language generation, translation, and comprehension.
\newblock \emph{arXiv preprint arXiv:1910.13461}.

\bibitem[{Lin(2004)}]{lin2004rouge}
Lin, C.-Y. 2004.
\newblock Rouge: A package for automatic evaluation of summaries.
\newblock In \emph{Text summarization branches out}, 74--81.

\bibitem[{Liu et~al.(2021)Liu, Liao, Wang, Zhang, Zhang, Liang, Wan, Li, Li,
  Zhang et~al.}]{liu2021medical}
Liu, G.; Liao, Y.; Wang, F.; Zhang, B.; Zhang, L.; Liang, X.; Wan, X.; Li, S.;
  Li, Z.; Zhang, S.; et~al. 2021.
\newblock Medical-vlbert: Medical visual language bert for covid-19 ct report
  generation with alternate learning.
\newblock \emph{IEEE Transactions on Neural Networks and Learning Systems},
  32(9): 3786--3797.

\bibitem[{Mrini et~al.(2021)Mrini, Dernoncourt, Chang, Farcas, and
  Nakashole}]{mrini2021joint}
Mrini, K.; Dernoncourt, F.; Chang, W.; Farcas, E.; and Nakashole, N. 2021.
\newblock Joint summarization-entailment optimization for consumer health
  question understanding.
\newblock In \emph{Proceedings of the Second Workshop on Natural Language
  Processing for Medical Conversations}, 58--65.

\bibitem[{Papineni et~al.(2002)Papineni, Roukos, Ward, and
  Zhu}]{papineni2002bleu}
Papineni, K.; Roukos, S.; Ward, T.; and Zhu, W.-J. 2002.
\newblock Bleu: a method for automatic evaluation of machine translation.
\newblock In \emph{Proceedings of the 40th annual meeting of the Association
  for Computational Linguistics}, 311--318.

\bibitem[{Qi et~al.(2020)Qi, Yan, Gong, Liu, Duan, Chen, Zhang, and
  Zhou}]{qi2020prophetnet}
Qi, W.; Yan, Y.; Gong, Y.; Liu, D.; Duan, N.; Chen, J.; Zhang, R.; and Zhou, M.
  2020.
\newblock Prophetnet: Predicting future n-gram for sequence-to-sequence
  pre-training.
\newblock \emph{arXiv preprint arXiv:2001.04063}.

\bibitem[{Radford et~al.(2021)Radford, Kim, Hallacy, Ramesh, Goh, Agarwal,
  Sastry, Askell, Mishkin, Clark et~al.}]{radford2021learning}
Radford, A.; Kim, J.~W.; Hallacy, C.; Ramesh, A.; Goh, G.; Agarwal, S.; Sastry,
  G.; Askell, A.; Mishkin, P.; Clark, J.; et~al. 2021.
\newblock Learning transferable visual models from natural language
  supervision.
\newblock In \emph{International conference on machine learning}, 8748--8763.
  PMLR.

\bibitem[{Thawkar et~al.(2023)Thawkar, Shaker, Mullappilly, Cholakkal, Anwer,
  Khan, Laaksonen, and Khan}]{thawkar2023xraygpt}
Thawkar, O.; Shaker, A.; Mullappilly, S.~S.; Cholakkal, H.; Anwer, R.~M.; Khan,
  S.; Laaksonen, J.; and Khan, F.~S. 2023.
\newblock Xraygpt: Chest radiographs summarization using medical
  vision-language models.
\newblock \emph{arXiv preprint arXiv:2306.07971}.

\bibitem[{Tiwari et~al.(2022)Tiwari, Manthena, Saha, Bhattacharyya, Dhar, and
  Tiwari}]{tiwari2022dr}
Tiwari, A.; Manthena, M.; Saha, S.; Bhattacharyya, P.; Dhar, M.; and Tiwari, S.
  2022.
\newblock Dr. can see: towards a multi-modal disease diagnosis virtual
  assistant.
\newblock In \emph{Proceedings of the 31st ACM international conference on
  information \& knowledge management}, 1935--1944.

\bibitem[{Yadav et~al.(2021)Yadav, Gupta, Abacha, and
  Demner-Fushman}]{yadav2021reinforcement}
Yadav, S.; Gupta, D.; Abacha, A.~B.; and Demner-Fushman, D. 2021.
\newblock Reinforcement learning for abstractive question summarization with
  question-aware semantic rewards.
\newblock \emph{arXiv preprint arXiv:2107.00176}.

\bibitem[{Zhang et~al.(2023)Zhang, Huang, Jin, and Lu}]{zhang2023vision}
Zhang, J.; Huang, J.; Jin, S.; and Lu, S. 2023.
\newblock Vision-language models for vision tasks: A survey.
\newblock \emph{arXiv preprint arXiv:2304.00685}.

\bibitem[{Zhang et~al.(2020)Zhang, Zhao, Saleh, and Liu}]{zhang2020pegasus}
Zhang, J.; Zhao, Y.; Saleh, M.; and Liu, P. 2020.
\newblock Pegasus: Pre-training with extracted gap-sentences for abstractive
  summarization.
\newblock In \emph{International Conference on Machine Learning}, 11328--11339.
  PMLR.

\bibitem[{Zhang et~al.(2019)Zhang, Kishore, Wu, Weinberger, and
  Artzi}]{zhang2019bertscore}
Zhang, T.; Kishore, V.; Wu, F.; Weinberger, K.~Q.; and Artzi, Y. 2019.
\newblock Bertscore: Evaluating text generation with bert.
\newblock \emph{arXiv preprint arXiv:1904.09675}.

\bibitem[{Zheng et~al.(2023)Zheng, Chiang, Sheng, Zhuang, Wu, Zhuang, Lin, Li,
  Li, Xing et~al.}]{zheng2023judging}
Zheng, L.; Chiang, W.-L.; Sheng, Y.; Zhuang, S.; Wu, Z.; Zhuang, Y.; Lin, Z.;
  Li, Z.; Li, D.; Xing, E.; et~al. 2023.
\newblock Judging LLM-as-a-judge with MT-Bench and Chatbot Arena.
\newblock \emph{arXiv preprint arXiv:2306.05685}.

\bibitem[{Zhou et~al.(2023)Zhou, He, Sun, Xu, Chen, Chu, Zhou, Liao, Zhang, and
  Gao}]{zhou2023skingpt}
Zhou, J.; He, X.; Sun, L.; Xu, J.; Chen, X.; Chu, Y.; Zhou, L.; Liao, X.;
  Zhang, B.; and Gao, X. 2023.
\newblock SkinGPT-4: An Interactive Dermatology Diagnostic System with Visual
  Large Language Model.

\bibitem[{Zhu et~al.(2023)Zhu, Chen, Shen, Li, and Elhoseiny}]{zhu2023minigpt}
Zhu, D.; Chen, J.; Shen, X.; Li, X.; and Elhoseiny, M. 2023.
\newblock Minigpt-4: Enhancing vision-language understanding with advanced
  large language models.
\newblock \emph{arXiv preprint arXiv:2304.10592}.

\end{thebibliography}

\end{document}


\maketitle

\section{Annotation Guidelines for creation of MMQS Dataset}

\begin{itemize}[label=--, left=0pt, labelsep=1em, itemsep=1em]

\item \textbf{Identification of Medical Disorder from Textual Symptoms:}Accurately identify the medical disorder based on the symptoms described in the patient's text.
  
\item \textbf{Infuse Multimodal context:}To infuse multimodal context information in the patient query, consider including statements like \textbf{\textit{'Doctor, could you please evaluate what's happening in my tonsils?'}} appropriately.
  
\item \textbf{Inclusion of Quality Disorder Image:}Add a high-quality image that corresponds to the identified disorder, aiding visual recognition.

\item \textbf{Ensuring Data Privacy:}Remove any personal information present to uphold data privacy standards.

\item \textbf{Reevaluation of Ambiguous Data Points:}Review cases where disorder prediction is uncertain; consult a medical professional for reassessment.

\item \textbf{Updating Golden Summaries with Visual Cues:}Fuse the textual symptom description with the identified disorder from the visual symbol in the old golden summary, creating a comprehensive updated golden summary.
\end{itemize}

\section{Data Samples for MMQS Dataset}
The complete dataset and code will be made available on successful acceptance of the paper.For now few more data samples from our MMQS dataset is given below for reference in Table-\ref{tab:questions_summaries}.The image folder is given seperately and images can be verified using the relative path.
\begin{table}[h]
    \centering
    \scalebox{0.85}{
    \begin{tabularx}{1.2\textwidth}{|X|X|X|}
        \hline
        \textbf{Question} & \textbf{Summary} & \textbf{Relative Image Path}  \\
        \hline
        11YR old daughter diagnosed MRSA+ 13 days ago started having high fever 5 days ago which we keep down with Advil cause also feeling unwell and severe fatigue. Noticed bad breath and very  with white spots all over them.Please doctor see the current condition below . Tonight goes to bed with fever and a bit of Advil. Later got rwally bad cold sweats cold body temp 94,9 and was shaking so kuch had to wake her was scaring me! Had to change her to dry pjs. Back to bed with again cold sweats and low body temp. Snoring. What should I do pls. & What could be the cause of high fever, swollen tonsils with white spots, severe fatigue, and cold sweats with low body temperature in an 11-year-old girl recently diagnosed with MRSA?&swollen tonsils/Image\_2.jpg \\
        \hline
        Is it safe for 2 year olds.My two year old has a bacterial infection in his tonsils and one of the medication prescribed was zyncet in addition to antibiotics.Please doctor see the current condition below .  Is Zyncet safe for a two year old. Also, his gums have recently become swollen after three days of use. Could this be a side effect of the drug. Kindly advise & Is Zyncet safe for a two year old with bacterial tonsil infection and can swollen gums be a side effect? &swollen tonsils/Image\_7.jpg \\
        \hline
        We have a 16 month old baby. She has been suffering from serious allergy for last 3/4 months. Her skin gets red as rash.Please doctor see the current condition below . It itches heavily. She feels discomfort in sleeping. Sleeping does not look like sound.Would you please suggest us what we should do now. It is mentioned here that I, the father of the baby have acute allergy from my age 23. & What treatment should be recommended for a 16-month-old baby with a persistent skin rash, itching, and disrupted sleep due to allergies? & skin rash/Image\_105.jpg \\
        \hline
        My duaghter 10 month old she have small swelling at right side of neck and his lab investigation has shown ESR 53mm/1st hr and Platelet 551 \%. pheripheral film shown anisocytosis hypochromatic poikilocytosis target cell platelet increase neutropenia.Please see something happened on neck below . & What is the significance of a neck swelling in a 10-month-old with abnormal lab results including elevated ESR and platelet count? & neck swelling/Image\_4.jpg \\
        \hline
        My sister has a really bad ulcer in her mouth.  Is there anything  that used to be sold by prescription that I can purchase at the drug store that can heal some issue in my mouths.Please doctor see the mouth below .  I need a medicine that formerly was sold by prescription as the over the counter medications don't seem to work.  She used to use Triamcinolone Acetonide Paste for mouth, but she is out of it. &What medication sold by prescription can be purchased over the counter to treat a severe mouth ulcer? & mouth ulcers/Image\_53.jpg \\
        \hline
        Hello, I have been taking cipralex for anxiety for about 6 weeks now, and lately I have noticed that my bottom lip is swollen - to the point that it is interfering a little with my speech and I also tend to bite into it when I am eating. My bottom lip droops down and has no bounce back when I pull it down. i am aware that I have to pull it back in to look normal .The image for that is attached below . Could this be caused by the cipralex. Should I stop taking the medication...I am out of town and will not see my doctor for another 4 days.- & Could lip swelling and drooping, potentially caused by cipralex, warrant immediate cessation of the medication before seeing a doctor? & lip swelling/Image\_1.jpg \\
        \hline
        Hi doctor,I am using hair dryer five days a week. Can I have brain tumor or cancer by using hair dryer. I have been using it for six months. My hairs are pretty well. I have a nice growth and I will oil my hair every day. Sometimes I have pimples in my scalp and there is a mild pain in my one side of the head which lasts for a few seconds.Please doctor see the eyes below  & What could be the cause of my sister's increasing skin rash, present for the last five years? & dry scalp/Image\_19.jpg \\
        \hline
        My lips and the inside of my jaw is sore and irritated slightly swollen I am a 36 years old black female, I work ine a medical facility. These symptoms started after wearing them same type of lipstick for two weeks color stay. I have worn it ine tje past burn not usually daily but thes symptoms have progressively gotten worst to then point where then corners of my mouth bleed. What could be causing this. I have DM, HBP, HI CHOLESTEROL.Please see the image of the affected area doctor  below  & What could be causing  sore, irritated and swollen lips and jaw, with bleeding corners of the mouth after using a specific lipstick? & knee swelling/Image\_2.jpg

    \end{tabularx}
    }
    \caption{Questions and Summaries}
    \label{tab:questions_summaries}

\end{table}